\documentclass[letterpaper, 10 pt, conference]{ieeeconf}  %

\IEEEoverridecommandlockouts                              %
                                                          
\usepackage[usenames,dvipsnames]{xcolor}

\usepackage{epsfig} %
\usepackage{graphics} %
\usepackage{amsmath} %
\usepackage{amssymb}  %
\usepackage{cite}	
\usepackage{graphicx}
\usepackage{tabularx}
\usepackage{lipsum}
\usepackage{multirow}
\usepackage{psfrag}
\usepackage{float}
\usepackage{placeins}

\usepackage{algorithm}
\usepackage{algorithmicx}
\usepackage[]{algpseudocode}

\usepackage{subfloat}
\usepackage{subfig}
\usepackage{caption}

\usepackage{soul}

\usepackage[colorlinks = true,
            linkcolor = black,
            urlcolor  = black,
            citecolor = black,
            anchorcolor = black]{hyperref}

\usepackage{bm}
\usepackage{microtype}
\usepackage{wasysym} %
\usepackage{adjustbox}
\usepackage{lipsum}

\usepackage{siunitx}
\DeclareSIUnit\gauss{G}

\usepackage{eurosym}

\usepackage{pgfplots}
\usetikzlibrary{shapes,arrows,fit,calc}
\usepackage[dvipsnames]{xcolor}
\usepackage{pgfplotstable}
\usetikzlibrary{math} %
\usepackage{tikz}
\pgfplotsset{compat=newest} 
\pgfplotsset{plot coordinates/math parser=false} 
\newlength\figureheight 
\newlength\figurewidth

\usepackage{array}
\newcolumntype{L}[1]{>{\raggedright\let\newline\\\arraybackslash\hspace{0pt}}m{#1}}
\newcolumntype{C}[1]{>{\centering\let\newline\\\arraybackslash\hspace{0pt}}m{#1}}
\newcolumntype{R}[1]{>{\raggedleft\let\newline\\\arraybackslash\hspace{0pt}}m{#1}}
\newcolumntype{M}{>{\centering\let\newline\\\arraybackslash\hspace{0pt}}X} %
\newcolumntype{s}{>{\hsize=.4\hsize}X}

\makeatletter
\newcommand{\thickhline}{%
    \noalign {\ifnum 0=`}\fi \hrule height 1pt
    \futurelet \reserved@a \@xhline
}
\newcolumntype{"}{@{\hskip\tabcolsep\vrule width 1pt\hskip\tabcolsep}}
\makeatother


\algnewcommand\algorithmicinput{\textbf{Initialization:}}
\algnewcommand\init{\item[\algorithmicinput]}

\algnewcommand\algorithmicalib{\textbf{Calibration:}}
\algnewcommand\calib{\item[\algorithmicalib]}

\algnewcommand\algorithmicevol{\textbf{Monitoring:}}
\algnewcommand\evol{\item[\algorithmicevol]}

\algdef{SE}[SUBALG]{Indent}{EndIndent}{}{\algorithmicend\ }%
\algtext*{Indent}
\algtext*{EndIndent}

\newcommand{\ie}{{i.e.}~}
\newcommand{\iec}{{i.e.},~}
\newcommand{\eg}{{e.g.}~}

\newcommand{\fig}{Figure~}

\newcommand{\tabb}{Table~}

\newcommand{\sect}{Section~}

\newcommand{\dofs}{DoFs~}

\newcommand{\TL}[1]{{\color{black} #1}}

\newcommand{\pg}[1]{\paragraph{\textbf{#1}}\mbox{}}

\definecolor{mycolor1}{rgb}{0.00000,0.44700,0.74100}%
\definecolor{mycolor2}{rgb}{0.85000,0.32500,0.09800}%
\definecolor{mycolor3}{rgb}{0.92900,0.69400,0.12500}
\definecolor{mycolor4}{rgb}{0.290,0.624,0.4}

\newcommand{\orcid}[1]{\href{https://orcid.org/#1}{\includegraphics[scale=0.5]{img/orcid_16x16.png}}}

\newcommand{\ND}[1]{{\color{black} #1}}

\title{\LARGE \bf 
Exploiting Intrinsic Kinematic Null Space for Supernumerary Robotic Limbs Control}

\author{T.~Lisini Baldi$^{1,2}$, %
		N.~D'Aurizio$^{1}$, %
		S.~Gurgone$^{3,4}$, %
		D.~Borzelli$^{3,5}$, %
		A. D'Avella$^{3,5}$, %
		and D.~Prattichizzo$^{1,2}$ %
\thanks{The research leading to these results has received funding from the European Union’s Horizon Europe programme under grant agreement No. 101070292 of the project \lq\lq HARIA - Human-Robot Sensorimotor Augmentation - Wearable Sensorimotor Interfaces and Supernumerary Robotic Limbs for Humans with Upper-limb Disabilities".}		
\thanks{$^1$ are with the Department of Information Engineering and Mathematics, University of Siena, Siena, Italy. 
}
\thanks{%
$^2$ are with the department of Humanoids and Human Centered Mechatronics (HHCM), Istituto Italiano di Tecnologia, Genova, Italy. %
}%
\thanks{$^3$ are with the Department of Biomedical and Dental Sciences and Morphofunctional Imaging, University of Messina, Messina, Italy.}
\thanks{$^4$ is with the Center for Information and Neural Networks, Advanced ICT Research Institute, National Institute of Information and Communications Technology, Osaka, Japan.}
\thanks{$^5$ are with the Laboratory of Neuromotor Physiology, IRCCS Fondazione Santa Lucia, Rome, Italy.}
}

\begin{document}

\maketitle

\begin{abstract}
Supernumerary robotic limbs (SRLs) gained increasing interest in the last years for their applicability as healthcare and assistive technologies. These devices can either support or augment human sensorimotor capabilities, allowing users to complete tasks that are more complex than those feasible for their natural limbs. However, for a successful coordination between natural and artificial limbs, intuitiveness of interaction and perception of autonomy are key enabling features, especially for people suffering from motor disorders and impairments. The development of suitable human-robot interfaces is thus fundamental to foster the adoption of SRLs.
 
With this work, we describe how to control an extra degree of freedom by taking advantage of what we defined the \textit{Intrinsic Kinematic Null Space}, \ie the redundancy of the human kinematic chain involved in the ongoing task. Obtained results demonstrated that the proposed control strategy is effective for performing complex tasks with a supernumerary robotic finger, and that practice improves users' control ability.
\end{abstract}

\setcounter{secnumdepth}{2}

\section{Introduction \label{SEC:intro}}
Supernumerary robotic limbs (SRLs) are wearable robotic devices designed to achieve human sensorimotor
augmentation \cite{raisamo2019human}.
Such devices enlarge the reachable workspace by adding artificial degrees of actuation to the
human body, giving the possibility of performing more complex actions with increased strength,
precision, and sensing capabilities. Differently from exoskeletons and exosuits, which are designed to mirror the kinematic structure of the body part on which they are worn and are used to empower human
natural movements, SRLs represent additional degrees of freedom (DoFs) that need to be controlled independently
from and/or simultaneously with biological limbs.

In less than a decade, we have seen the development of SRLs with different usages (fingers, hands,
arms, legs), actuation systems (fully actuated, underactuated), and design features (rigid/soft materials, level
of anthropomorphism, etc) \cite{prattichizzo2021human}. 
Even if these devices are not naturally part of the body, an appropriate human-robot interface can allow to perceive them as a real extension of the user. Rossi et al. \cite{rossi2021emerging} discovered the emerging of new bioartificial corticospinal motor synergies using a robotic additional thumb, which suggests how our motor system is open to very quickly embody the supernumerary robotic finger into the user’s body schema. 

\begin{figure}
\centering
\includegraphics[width=.85\columnwidth]{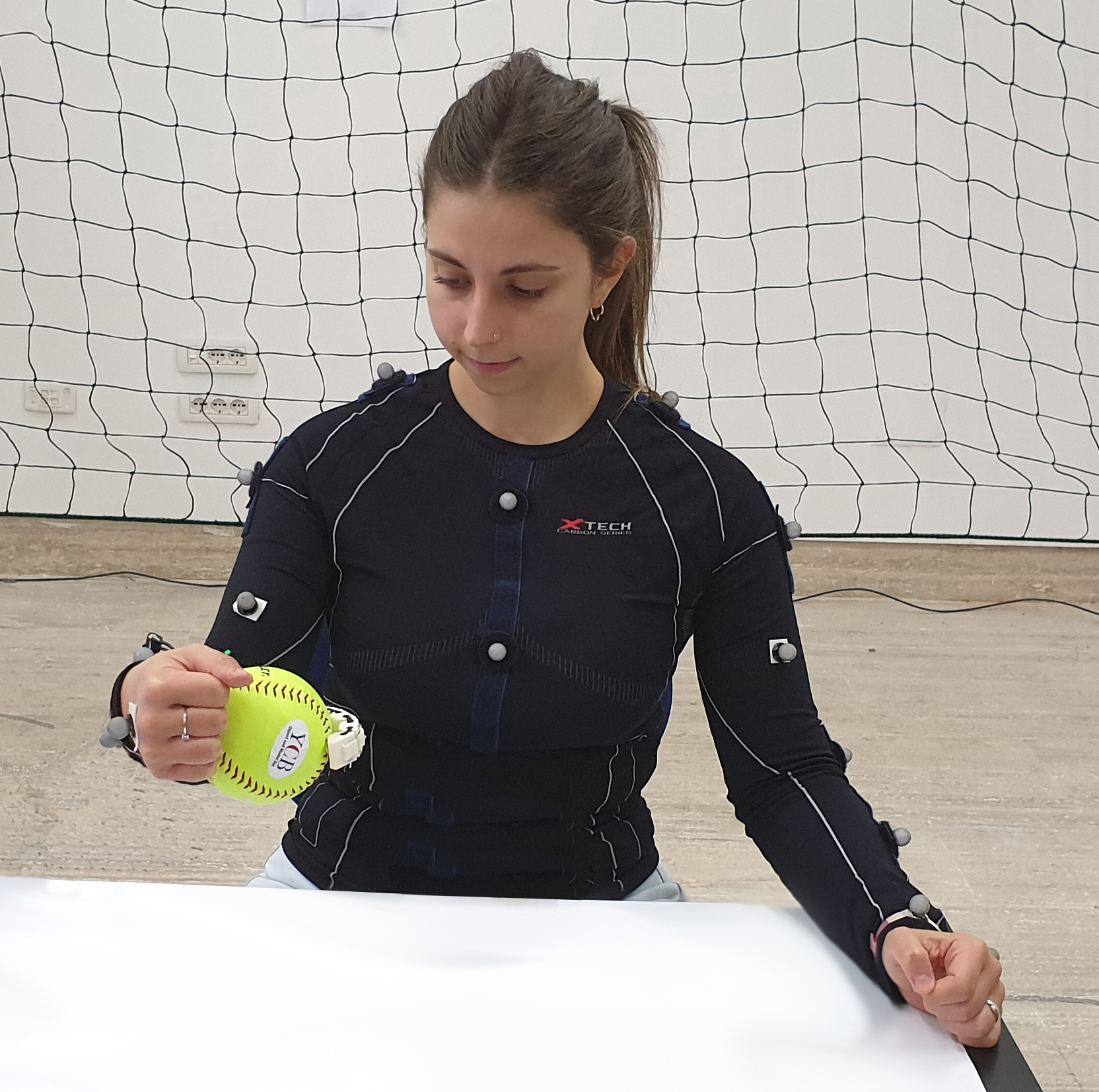}

\caption{A subject exploiting her Intrinsic Kinematic Null Space for controlling the supernumerary robotic finger.\label{FIG:concept}}
\end{figure} 

If adopted as technological support for service and assistance to people with disabilities, ease of use and ease of learning become fundamental requirements for guaranteeing device acceptance. %
For this class of users it is even more important to boost the user-device synergy and, consequently, reduce as much as possible the need of help to manage daily living.

The more intuitive is the control strategy, the faster is the embodiment of the supernumerary limb \cite{umezawa2022bodily}. Functional and user-based control techniques are thus key factors for encouraging users to take advantage of this technology rather than perceiving it as a source of frustration. From the user's point of view, it is possible to distinguish two different control strategies: \textit{non-autonomous} and \textit{autonomous}. 
In this paper we will focus on \textit{non-autonomous} control, which refers to an intentional and dedicated control with detailed instructions from the user. 
For instance, the user opens or closes a supernumerary robotic finger by pressing a push button embedded on a ring worn on the healthy hand as in \cite{hussain2017soft}. With \textit{autonomous} control, we indicate instead a control signal which adapts to the user actions without receiving explicit instructions. As an example, the exoskeleton supporting the wearer's gait should follow the intention to walk without waiting for specific commands on when to take a step.

As regards the \textit{non-autonomous} control of SRLs, 
the design and the purpose of these devices has a tight connection with the implemented control strategy. On the one hand, it is not surprising that the higher is the number of sensors and actuators in the device, the more accurately it can be moved and may follow the desired trajectory. On the other hand, the presence of multiple actuators implies a larger robot operational and configuration space, thus need the implementation of high level control strategies that allow users to control the coordinated motion of the degrees of freedom of the limb, and not the single motor \cite{prattichizzo2014object,legrand2022simultaneous}. 
Conversely, supernumerary robotic limbs with a low number of degrees of freedom are easier to control from the user point of view, exploiting, for example, wearable interfaces like buttons (e.g. ring interfaces \cite{hussain2015vibrotactile, hussain2017soft, hussain2019design}). As a final consideration, since SRLs are designed to cooperate with the human’s natural limbs, the major challenge in the development of an effective control strategy is guaranteeing that it does not affect the motion of the natural limb. In other words, control policies should allow an unobtrusive collaboration between robotic and biological limbs.

In this context, the purpose of this research is to present a new control paradigm that is strongly centered on users and their tasks. The fundamental concept is to capitalize on the human musculoskeletal system's redundancy to manage additional DoFs. While here we target the kinematic redundancy, \ie motions of the body that do not affect the action performed by the natural limbs, results on using the muscular redundancy, that is muscle activation patterns that do not produce net joint torques (such as the co-contraction of two antagonistic muscles that balance each other out), are already known and available in \cite{gurgone2022simultaneous}.

With this work, the \textit{Intrinsic Kinematic Null Space} control paradigm is defined and applied to control a supernumerary robotic limb, more specifically, a wearable extra-finger. To understand the potential effectiveness of this approach, the reader should consider the several movements that can be performed with the task to be accomplished being equal. Indeed, given the complexity of the human body, having redundancy while accomplishing a task is not an exception since the human body has about 244 degrees of freedom (thanks to 148 bones moved by 147 joints)~\cite{zatsiorsky1998kinematics}. 
\TL{
Instead of a model-based approach, we decided to pursue a data-driven method. Implementing a one-size-fits-all approach by identifying, calculating, and generalizing the null space of the Jacobian matrix for every task does not adequately address the specific constraints faced by people with disabilities.
A more tailored approach is necessary to account for individual needs and abilities.} To the best of our knowledge, this represents the first attempt to investigate the feasibility and usability of this novel control strategy for \mbox{human-device} interaction (\fig\ref{FIG:concept}).

\section{Intrinsic Kinematic Null Space Control}
\subsection{Definition\label{SUBSECT:definition}}

Considering the kinematic space of the whole human body, the kinematic null space is defined by the set of joint velocities that do not produce any hand (or foot) velocity in the given configuration of the natural limb. Once the task to be accomplished is identified, we distinguish between:
\begin{itemize}
\item[\textit{i)}] Extrinsic Kinematic Null Space (EKNS);
\item[\textit{ii)}] Intrinsic Kinematic Null Space (IKNS).
\end{itemize}
The kinematic null space of joints which are not involved in such a task belongs to the first category, whereas the IKNS refers only to joints directly employed in the task. For instance, grabbing a box with two hands directly involves joints of shoulders, upper arms, forearms, and wrists, which are all considered for the IKNS computation. The kinematic null space of all the other joints (e.g., knees and ankles) is the EKNS.

\subsection{Identification\label{SUBSECT:identification}}

Ages, habits, and lifestyles strongly influence how people interact with objects and surroundings, making individuals prone to perform tasks differently. Hence, calculating a-priori the kinematic intrinsic null space on the basis of existing kinematic models would have signified discarding the inter-individual variability, which instead is one of the strengths of this approach. For this reason we developed a procedure to identify the subjective kinematic intrinsic null space. Even if the proposed methodology can be generalized to control an arbitrary number of supernumerary robotic limb DoFs in various tasks, in this work we focused on the case of a single-arm task and we tested the effectiveness of the control strategy for one DoF only.

\pg{Kinematic chain identification} The first step of the procedure comes from the intrinsic kinematic null space definition (see \sect \ref{SUBSECT:definition}). In other words, the kinematic chain involved in the task at hand need to be a-priori identified, and this can be done by considering some general knowledge on the kinematic constraints of the human body.

For instance, considering a single-arm task, the hand represents the end-effector, and consequently the joints of shoulder, elbow, and wrist, and the associated links are the significant chain for the IKNS computation. User’s arm joint velocity vectors which do not contribute to change the hand velocity are then considered as belonging to the IKNS.

\pg{Joints motion analysis} Once the task and the significant chain have been clearly defined, the user is asked to perform the task, and its movements are recorded, analysed, and reconstructed using a motion capture system to estimate the joint values. Joint velocity vectors are considered as belonging to the IKNS if their contribution to the end-effector velocity is negligible for the particular application. In the case of single-arm task, the hand speed is assumed \TL{null} when its norm is lower than \SI{0.05}{\m\per\s}. This threshold is defined depending on the task.

\pg{Principal Component Analysis} To project the multidimensional space of the IKNS in the extra DoFs space, the acquired values are processed through Principal Component Analysis (PCA) and reduced into a set of values of linearly uncorrelated variables, \ie the Principal Components (PCs). 

To deal with the case of a single DoF, the 
\TL{normalized weighted vector sum}  of the extracted principal components that explain at least 80\% of the total variation is taken as direction for controlling the intended DoF and denoted with $Z \in \mathcal{R}^{1\times n}$, where $n$ is the cardinality of the IKNS.

\begin{figure}[t]
\centering\
\subfloat[]{\label{FIG:trajectory}\includegraphics[width=.5\columnwidth]{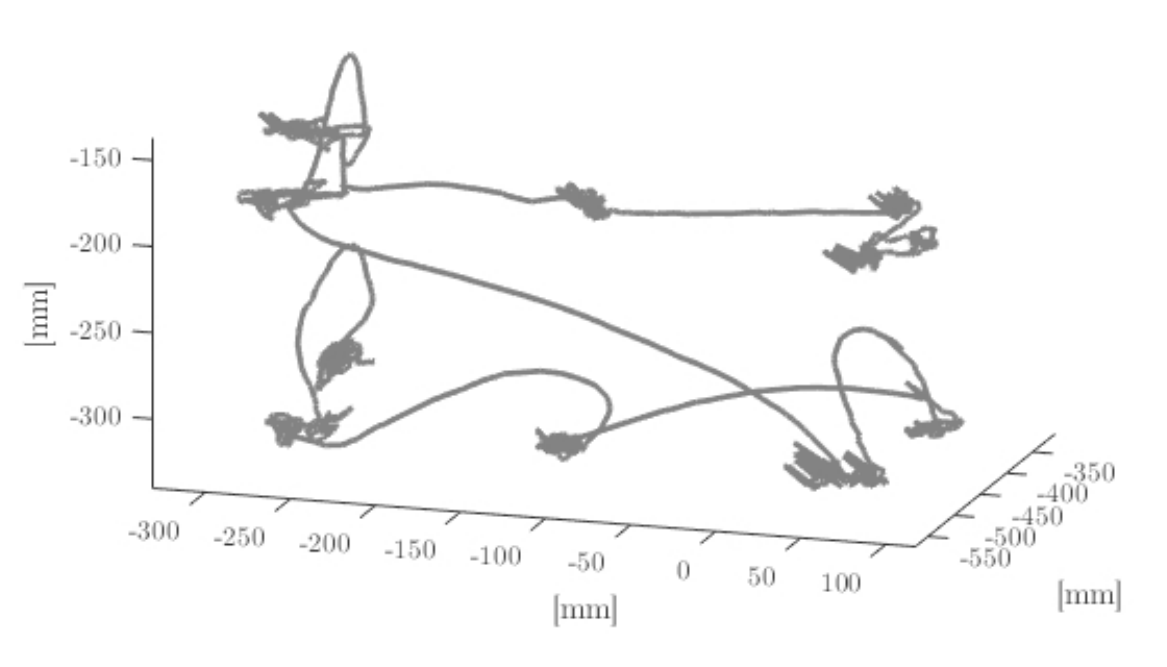}} %
\subfloat[]{\label{FIG:cluster2}\includegraphics[width=0.5\columnwidth]{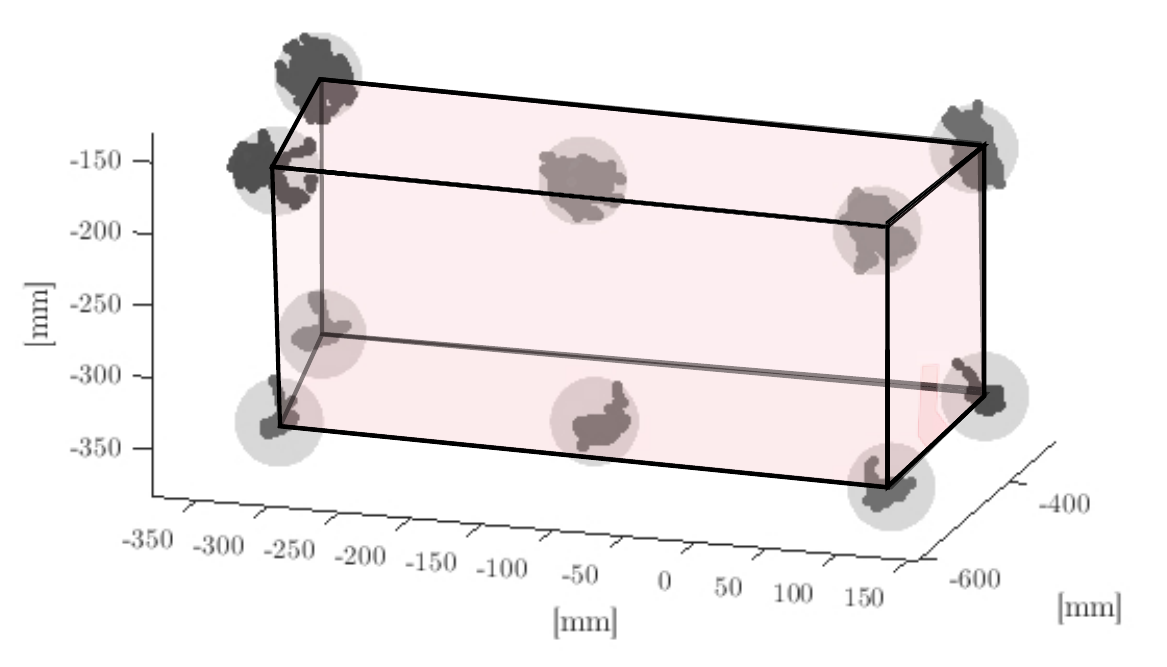}}
\caption{From data collection to interpolation volume estimation in a representative trial. In (a), the trajectory depicted by the marker attached to the user's right hand. In (b), data assigned to clusters (depicted with grey spheres) and considered for the PCA. The interpolation volume is depicted in light red.
\label{fig:clusterAll}}
\end{figure}

\pg{Workspace clustering} Since the IKNS varies at different positions of the end-effector, in theory computing the null space (and consequently the PCs) for each point in the arm workspace should be necessary to have a complete IKNS evaluation. This approach being not practicable and time consuming, we eased the process computing the null space base in any point of the user’s workspace starting from its value in a limited and predefined set of points where the user motions are recorded, referred to as ‘PCA points’. The latter are chosen to cover the dexterous region of the arm workspace and discard the boundaries, \ie the region of the reachable workspace where the mobility of the arm is reduced. For instance, when the arm is fully extended, the subject cannot move the arm arbitrarily without moving the hand, thus acquiring data in such kinematic configuration is useless. 

Data captured in the PCA points are clustered using an algorithm based on the k-means approach which implements the following steps:
\begin{enumerate}
\item compute a minimal bounding box for the recorded data;
\item initialize the N centroids (i.e.,one for each PCA point) on the surface of the bounding box;
\item for each data point x, compute the euclidean distance D(x) between x and each centroid, and assign each observation to the cluster with the closest centroid;
\item compute the new centroid locations as the average of the observations in each cluster;
\item repeat steps (3) through (5) until cluster assignments remain unchanged, or the maximum number of iterations is reached.
\end{enumerate}
Once the clusters are obtained, data at a distance greater than an appropriate threshold (evaluated and refined in accordance with the task characterization) from the centroids are discarded. Remaining data are used to compute and store $Z$ for each cluster, as well as minimum ($m$) and maximum ($M$) values of the user motion along $Z$ for normalization purposes. 
\TL{ $Z \in \mathbb{R}^{1\times n}$ is the direction that projects the current $n$-dimensional joint vector $\mathbf{q}$ into a monodimensional space.}

In the considered case of a single-arm task, the IKNS changes depending on the relative position of the hand with respect to the chest. Hence, the null space computed when the hand is near the chest is different from the one computed when the arm is outstretched ahead.

\pg{Workspace interpolation}
Finally, the PCs evaluated in each PCA point are used to create an interpolation volume and compute the IKNS-based control signal in any point of the dexterous workspace, also in those points in which the null space has not been recorded. 
Given the interpolation volume, a 3D Delaunay triangulation-based natural neighbour interpolation \cite{lee1980two,cazals2004delaunay} is used to reconstruct \ND{the proper direction. Thanks to this method we can compute}
 online and seamless the direction associated to the current null space as a smooth approximation 
of the directions $Z$ of the nearest clusters. Thus, the control signal $c$ is calculated as:
\begin{equation*}
c = \frac{\hat{Z}\mathbf{{q}}-\hat{m}}{|\hat{M}-\hat{m}|}
\end{equation*}
where $\hat{Z} \in \mathbb{R}^{1\times n}$ is the interpolated direction that projects the current $n$-dimensional joint vector $\mathbf{q}$ into the monodimensional space of DoF, while $\hat{m}$ and $\hat{M}$ result from the interpolation of $m$ and $M$, and are used to normalize the the control signal in the range from 0 to 1. Outside the interpolation volume, $Z$, $m$ and $M$ of the nearest cluster are taken to compute the value for controlling the DoF.

\fig\ref{fig:clusterAll} shows the steps from data collection to the estimation of the interpolation volume. The trajectory of the hand is in \fig\ref{FIG:trajectory}, while \fig\ref{FIG:cluster2} reports the identified clusters highlighted with grey spheres, and the interpolation volume highlighted with a light red solid. Data laying outside the spheres have been discarded.

\section{Experimental Campaign\label{SEC:exp}}
We evalutated the IKNS control paradigm described above with an experimental campaign. The aim was to prove that the proposed methodology is a viable framework for controlling a supernumerary robotic finger in an effective way, investigating also the learning process of the users. Experiments were conducted both in virtual and in real environments, and designed to meet the final usage of the supernumerary robotic finger in activities of daily living (ADLs): grasp objects by opening/closing the device without moving the hand.

Each participant gave their written informed consent to participate and was able to discontinue participation at any time during the experiments. The experimental evaluation protocols followed the declaration of Helsinki, and there was no risk of harmful effects on participants' health. Data were recorded in conformity with the European General Data Protection Regulation 2016/679, stored  on local repositories with anonymized identities (i.e., User1, User2), and used only for the post processing evaluation procedure. Please note that no sensible data were recorded.%

Ten subjects were enrolled in the experimental campaign (seven males and three females, from 22 to 57 years old, mean $35\pm4.5$, all right-handed). None of them had previous experiences in controlling wearable robots. Each subject started the experimental session with a calibration procedure, followed by two experiments with a resting period of more than an hour between them. A training phase of five minutes was provided at the beginning of each experiment to acquaint the participants with the system. 

Experiments were carried out in a room equipped with a Vicon tracking system with ten Vicon Bonita cameras. The subject was positioned at the centre of the room with retro-reflective markers attached to different location of its upper body to track arm joint values. Cameras were positioned at the upper corners of the room (two per corner with different orientations, for a total of eight cameras), and on the left and right side of the subject, fixed to tripods placed on opposite sides of the room. The body posture was reconstructed online with a frame rate of \SI{100}{\Hz} using Vicon Nexus Software v3.10 (Vicon Motion Systems Ltd, UK).

\begin{figure*}
	\centering
	\null \hfill %
	\subfloat{
		\includegraphics[width=0.55\columnwidth]{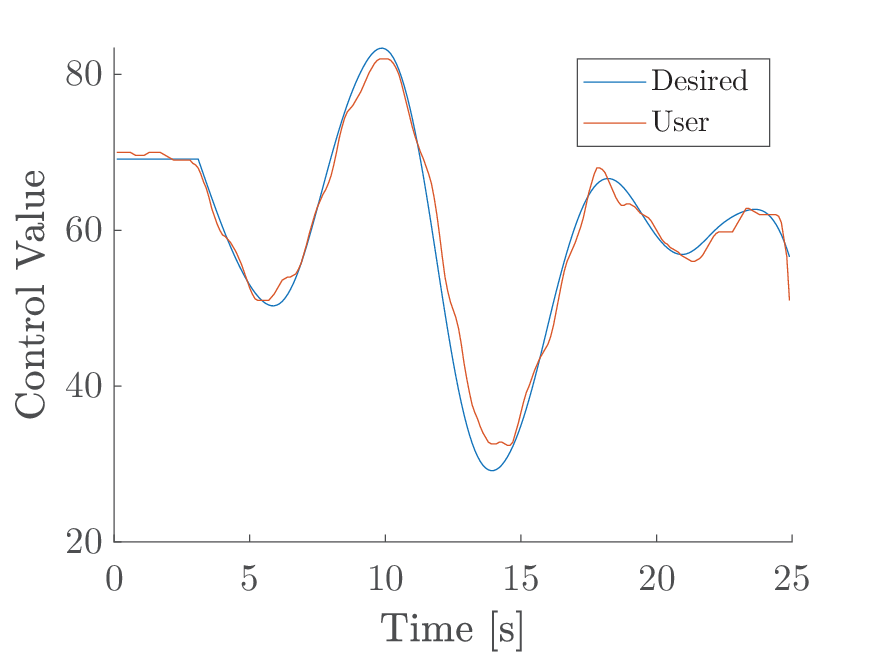}
	} \hfill %
	\subfloat{
		\includegraphics[width=0.55\columnwidth]{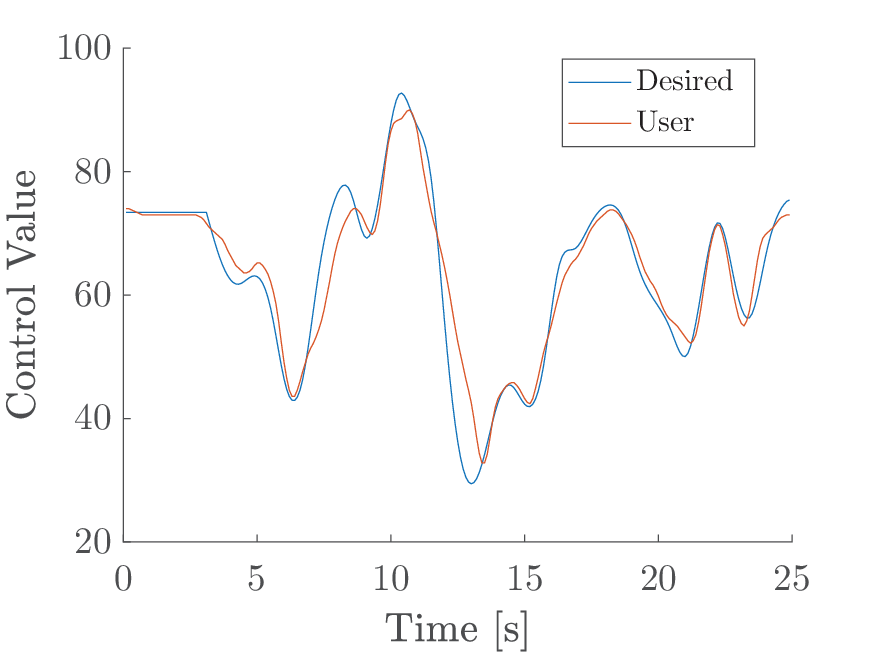}
	}\hfill %
	\subfloat{
		\includegraphics[width=0.55\columnwidth]{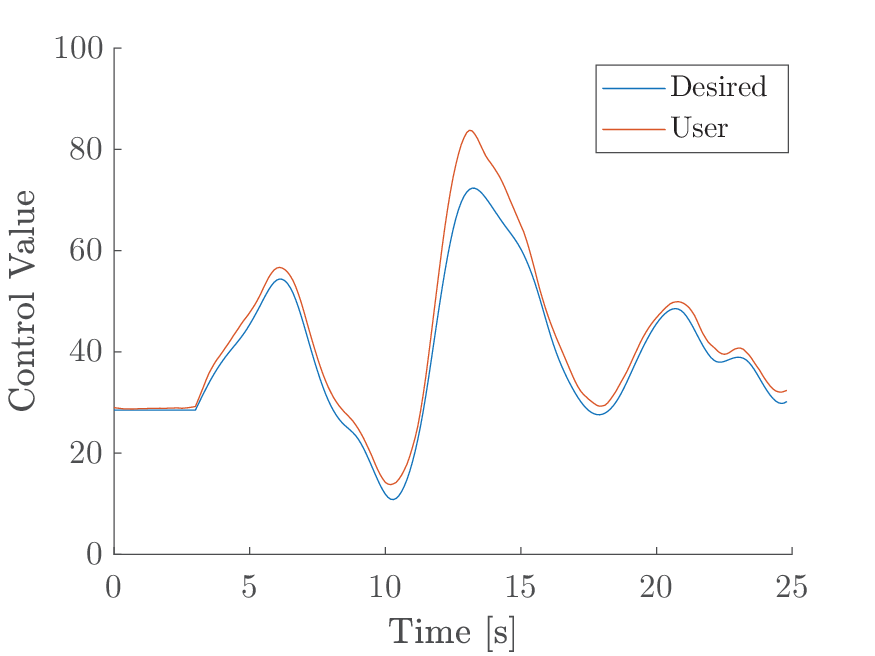}
	}\hfill \null \\ \vspace{-15px}
	\null \hfill %
	\setcounter{subfigure}{0}%
	\subfloat[Trajectory 1]{
		\includegraphics[width=0.55\columnwidth]{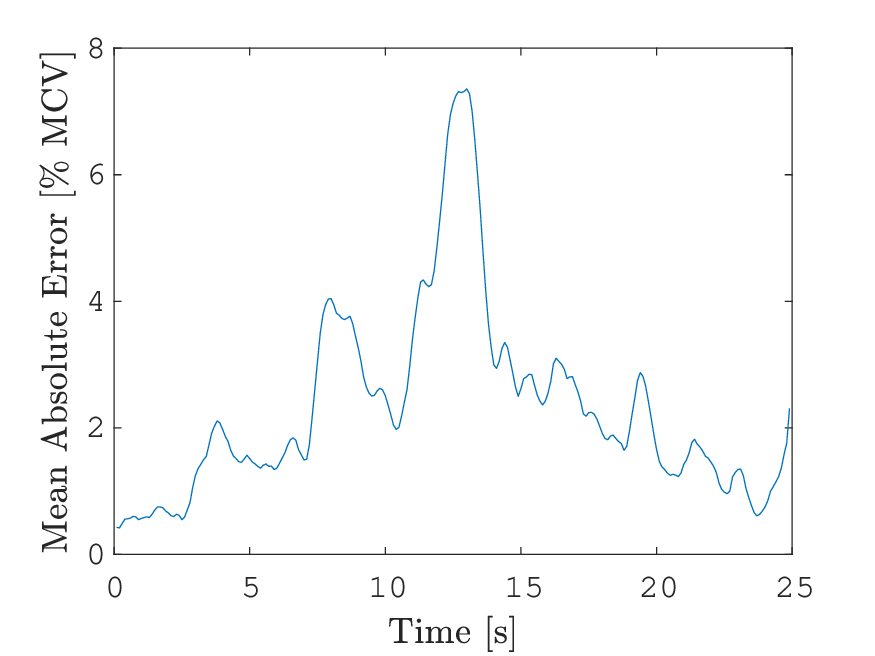}
		\label{Fig:traiettoria1}
	} \hfill %
	\subfloat[Trajectory 2]{
		\includegraphics[width=0.55\columnwidth]{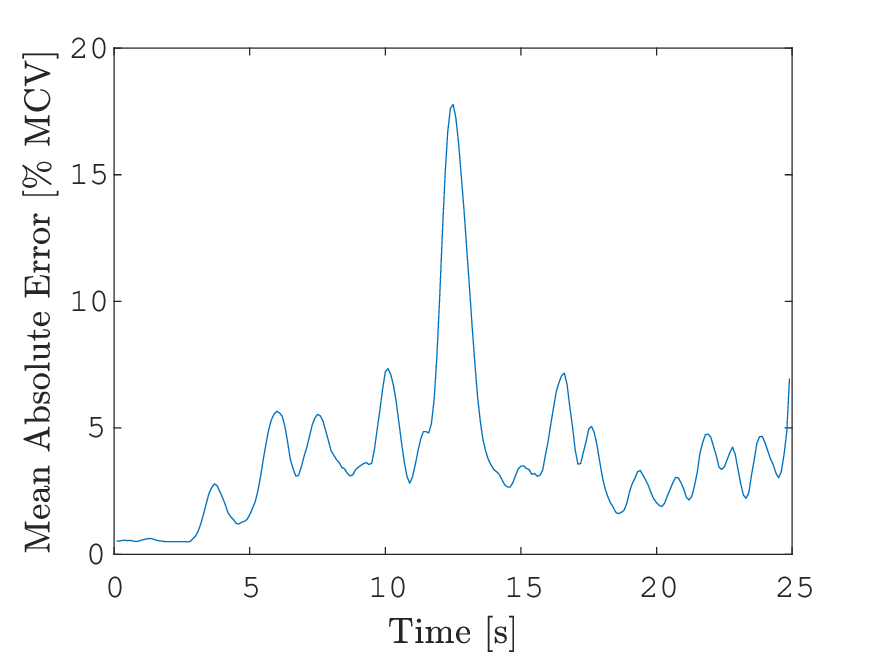}
		\label{Fig:traiettoria2}
	}\hfill %
	\subfloat[Trajectory 3]{
		\includegraphics[width=0.55\columnwidth]{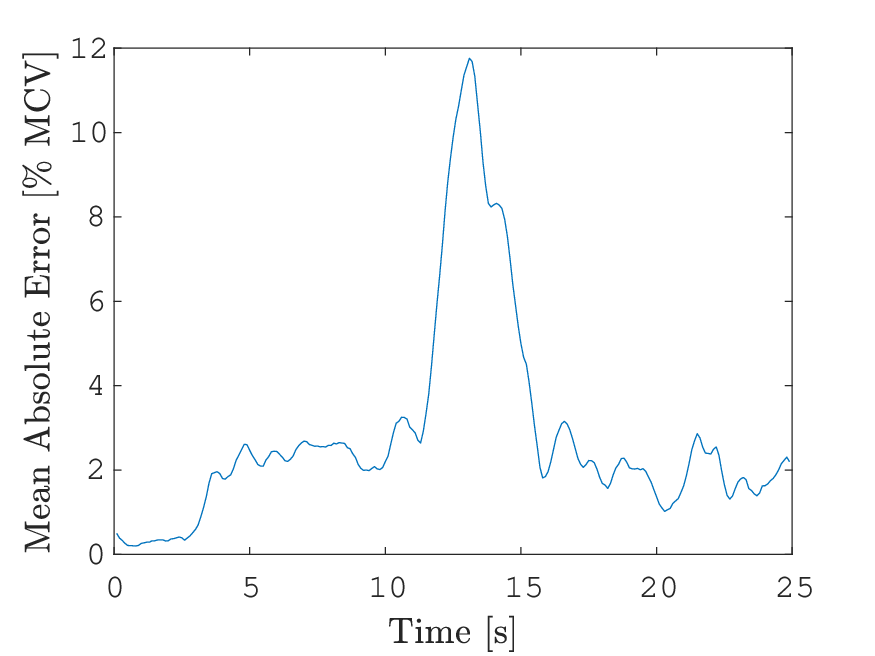}				
		} 
	\hfill  \null
\caption{Results of the experiment conducted in virtual environment. Upper panels: desired trajectories (blue line) together with the trajectory of the pointer controlled by the user in a representative trial (red line).
Lower panels: mean absolute error (expressed as percentage of Maximum Control Value) computed among all the participants for each desired trajectory. It is worth noticing that the average absolute error among all the trials is lower than 4.5 for a control value ranging from 0 to 100. Lower panels make evident that the absolute error is lower than 4	when the desired profile is smoothly changing, while reaches greater values when the user is requested to follow rapid changes.}\label{FIG:Exp2_profiles}
\end{figure*}

\subsection{Calibration}
\label{SEC:calibration}
24 retro-reflective markers were attached to the subject in accordance with 
the Oxford Upper Body Model \cite{nexusUpper}. %
To calibrate the system, each participant was asked to stand at the centre of the room for \SI{5}{\s}. A static acquisition and anthropometric measurements were used to create each user's upper body skeleton model, consisting in 24~\dofs.%

Thanks to the calibration procedure, the user's skeleton is automatically reconstructed and the kinematic model is fitted online. %
This step enables the real time capturing of joint angle values and body segments positions.
To gain awareness of the workspace, participants were asked to seat and explore the arm workspace with the hand without moving the torso. %
After one minute of free exploration, participants were asked to visualize an imaginary parallelepiped covering their arm workspace and %
were instructed to place the hand in 10 points (8 in the proximity of the vertexes and 2 at the centres of the upper and lower surfaces) %
and freely move the arm for five seconds, holding the hand as steady as possible (\iec without changing position and orientation). They were asked to explore the entire range of motion available in each position, so as to record minimum and maximum reachable values.
The algorithm described in \sect\ref{SUBSECT:identification} selected meaningful data and computed the Principal Components for each of the ten points and the control signal in any point of the working space, depending on the posture of the user. %

\subsection{Virtual Environment}
The first experiment aimed at answering the following research question:
\textit{Is the generated control signal appropriate for attaining a desired behaviour?}
We answered by evaluating the user's accuracy in a tracking task, \iec in following a predefined reference profile exploiting the proposed control system. %
This kind of setup was thought to  simulate the execution of precise opening/closing movements of the supernumerary robotic finger using the IKNS control.

Three desired trajectories \ND{for the control signal $c$} (depicted in \fig\ref{FIG:Exp2_profiles} and denoted with T1, T2, and T3 in what follows) were pseudo-randomly generated.
The user was seated in front of a screen and was 
 instructed to follow the displayed predefined profiles. 
 This had to be done by using arm movements belonging to the IKNS to control the vertical displacement of a circular red pointer, while  the horizontal  displacement was updated at constant  velocity. %
 Control values were normalized so as to range from 0 to 100.
Each profile was  repeated three times, for a total of nine trials per each subject. %
A time of five seconds  was provided to reach the starting condition (i.e., to align the position of the  pointer with the initial flat trend of the trajectory, see \fig\ref{FIG:Exp2_profiles}), then a pop-up window informed the user about the starting of the experiment. 
During the trial, a red line joined the positions already reached by the pointer and informed the user about the progress.
Each trajectory lasted \SI{25}{\s}, thus the total time of each trial was \SI{30}{\s}.

For the sake of comparison, we asked each subject to repeat the same experiment using a commercial gamepad (F310, Logitech, CH), being this device a gold standard and hence suitable as a control condition for interpreting the results.

\pg{Metrics of interest}
We considered the performance in following a desired profile as metric of success in accomplishing the task. %
Similarly to \cite{parietti2017independent}, for each trial we defined the  Root Mean Square Error (RMSE) as \mbox{$RMSE_t = \sqrt{ \frac{1}{N}\sum_{i=1}^{N}(y_{t,i} - y_i)^2}$}, where N is the number of samples in a trial, $y_i$ is the actual control value, and $y_{t,i}$ is the corresponding target value. %
Notice that the tracking RMSE is a suitable metric to evaluate the rapidity and the accuracy of the robot motions \cite{krakauer2011human}. This is due to the fact that the RMSE increases both if users are slow in adapting the control variable and if they miss the targets. %
In other words, human control has to be simultaneously fast and accurate to yield a low RMSE. %

\pg{Results}
Outcomes show a small RMSE in performing the experiment using the IKNS control. The mean error among the participants was $3.35 \pm 0.83$, 
$5.43 \pm 0.92$, and $4.50 \pm 1.54$ for T1, T2, and T3, respectively. In all cases it was lower than 5.5, which represents the 5.5\% of the maximum control value as the control signal ranges from 0 to 100. These values are comparable with those obtained using the gamepad, \ie $4.13 \pm 0.76$ for T1, $4.98 \pm 1.33$ for T2, and $4.15 \pm 1.10$ for T3. In \fig\ref{FIG:Exp2_profiles} we report a representative trial (upper panel) and the mean RMSE among participants (lower panel) for each trajectory. 

A statistical analysis was conducted to compare results obtained with the two control paradigms. 
A paired-samples t-test revealed that there is no statistically
significant difference between the mean RMSEs recorded with the IKNS control and those recorded with the gamepad control 
($p = 0.979$, $p > 0.05$, $t(29) = 0.27$). No outliers were detected and the assumption of normality was not violated, as assessed by Shapiro-Wilk's test ($p = 0.352$, $p>0.05$). 

The fact that there were no statistically significant differences between the two control techniques suggests that the IKNS control is intuitive enough to allow users to exploit it with good skills from the first approach, with performance comparable to those obtained with the most widespread and used controller. In addition, a careful examination of the lower panels of \fig\ref{FIG:Exp2_profiles} makes evident that the absolute error is lower than 4 when the desired profile is smoothly changing, while reaches greater values when the user is requested to follow rapid changes of the path.  With both the control strategies, the highest error derives from the second trajectory (\fig\ref{Fig:traiettoria2}), which requires the user to rapidly modify the control value moving from 90 to about 30 in less than 3 seconds.  %
Conversely, the first trajectory (\fig\ref{Fig:traiettoria1}) requires the user to perform slower movements, and the resulting mean error for the IKNS control is less than 8 for each participant. %

\subsection{Real Environment}
\label{SEC:ExpReal}
On the basis of the promising results obtained in the previous experiment, we proceeded to evaluate the effectiveness of our system in a real scenario. %
More in detail, the second experiment aimed at tackling the following research question: \textit{Is the proposed approach suitable to accomplish common activities of daily living?}
To answer, we asked users to perform a pick-and-place task with multiple objects and different target locations.
Users wore the supernumerary robotic finger on the right forearm, mimicking a post-stroke hemiparesis, and controlled the opening/closing actuation with the same arm using the IKNS control strategy. The supernumerary robotic finger utilised for the experiment is a modified version of the one presented in \cite{PrMaHuSa-roman14}. It is worth noticing that, to demonstrate the capability of the proposed approach, the degree of actuation of the wearable device was controlled in continuous manner, and the range of the IKNS control signal was mapped into the finger range of motion.

\begin{figure}[t]
\centering\
\null \hfill
\subfloat[]{\label{FIG:exp2_1}\includegraphics[trim={30cm 15cm  35cm 45cm}, clip, width=0.47\columnwidth]{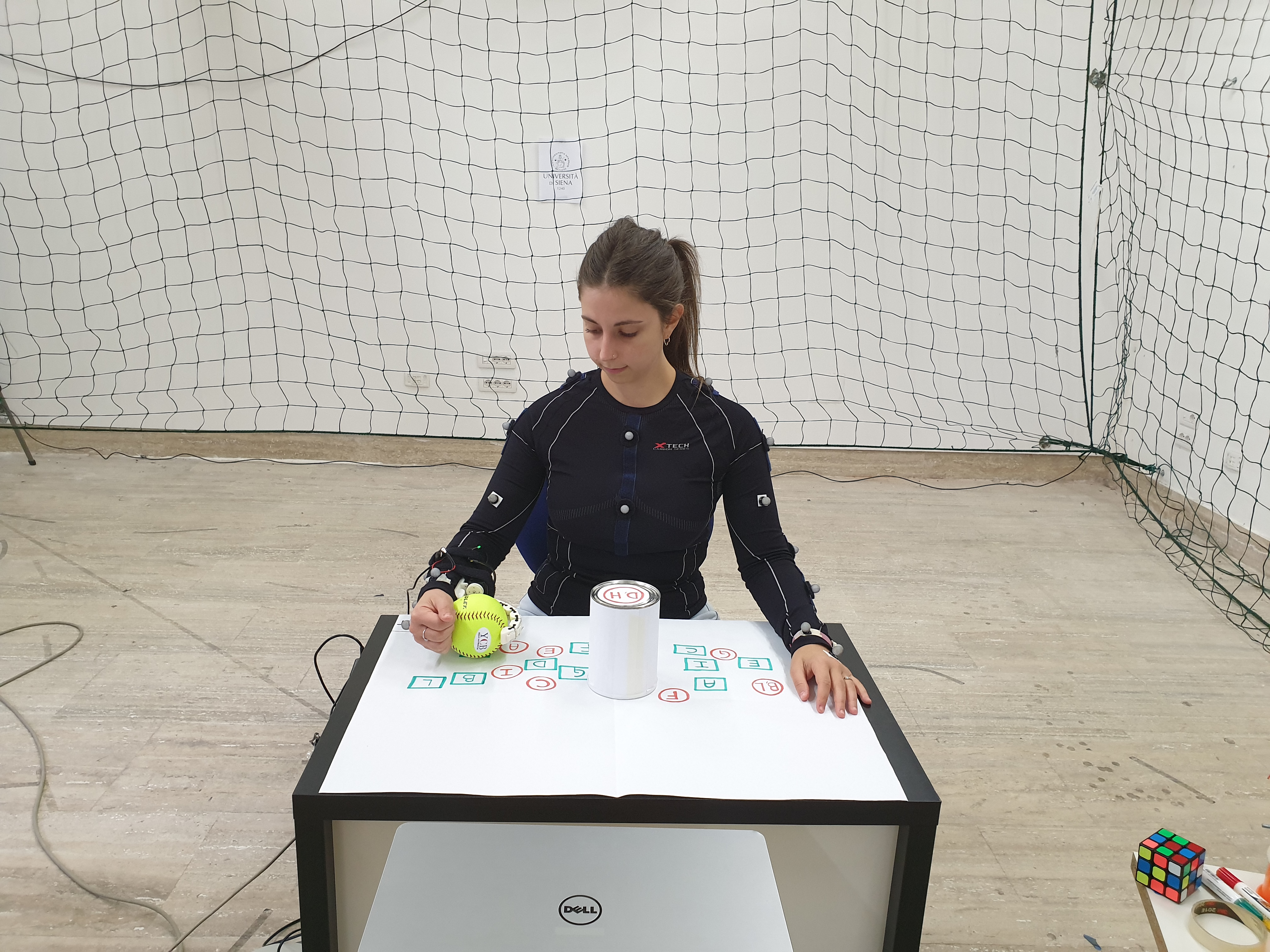}}
\hfill
\subfloat[]{\label{FIG:exp2_2}\includegraphics[trim={30cm 15cm  35cm 45cm}, clip,width=0.47\columnwidth]{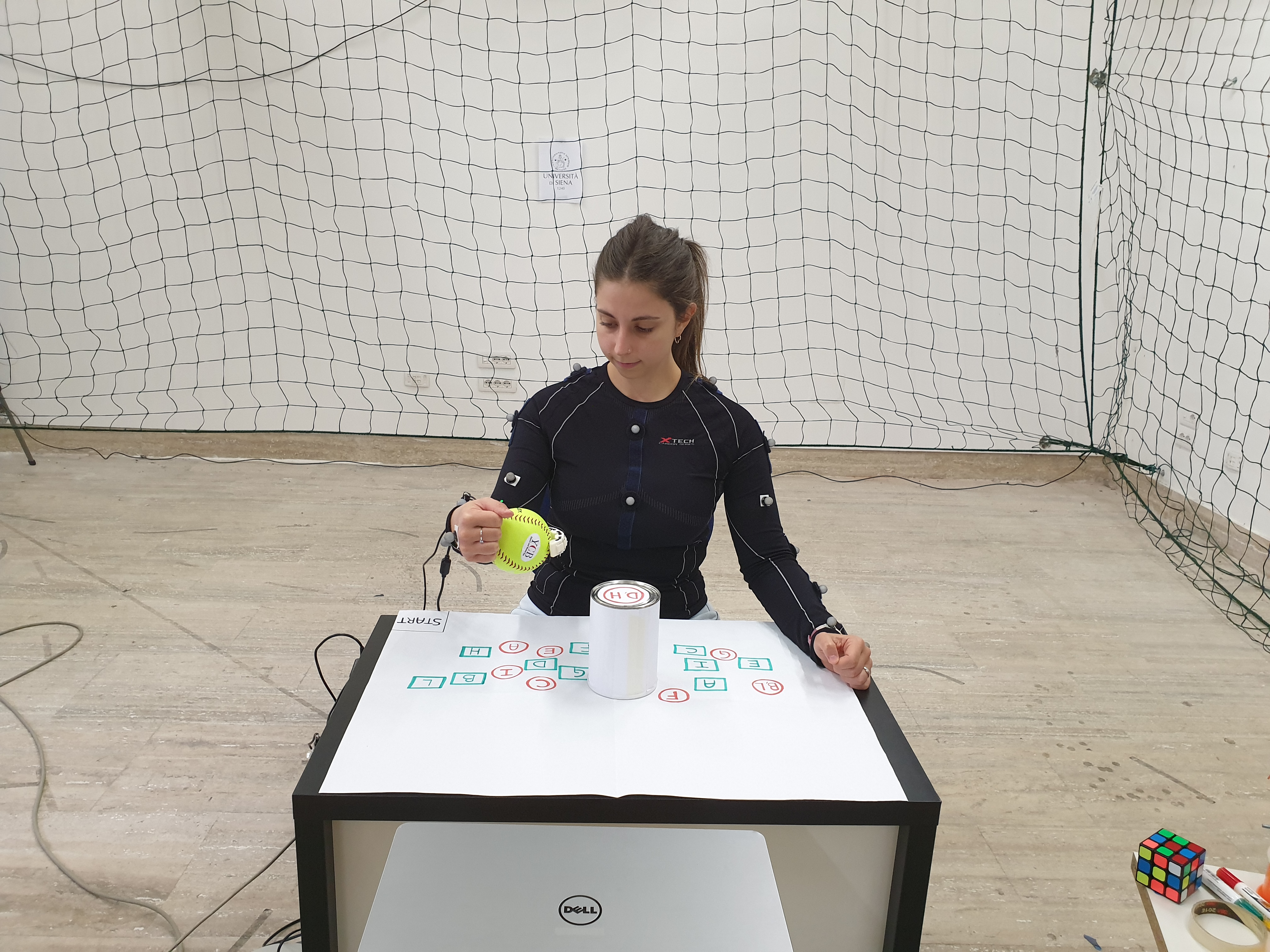}}
\hfill \null \\
\null \hfill
\subfloat[]{\label{FIG:exp2_3}\includegraphics[trim={30cm 15cm  35cm 45cm}, clip,width=0.47\columnwidth]{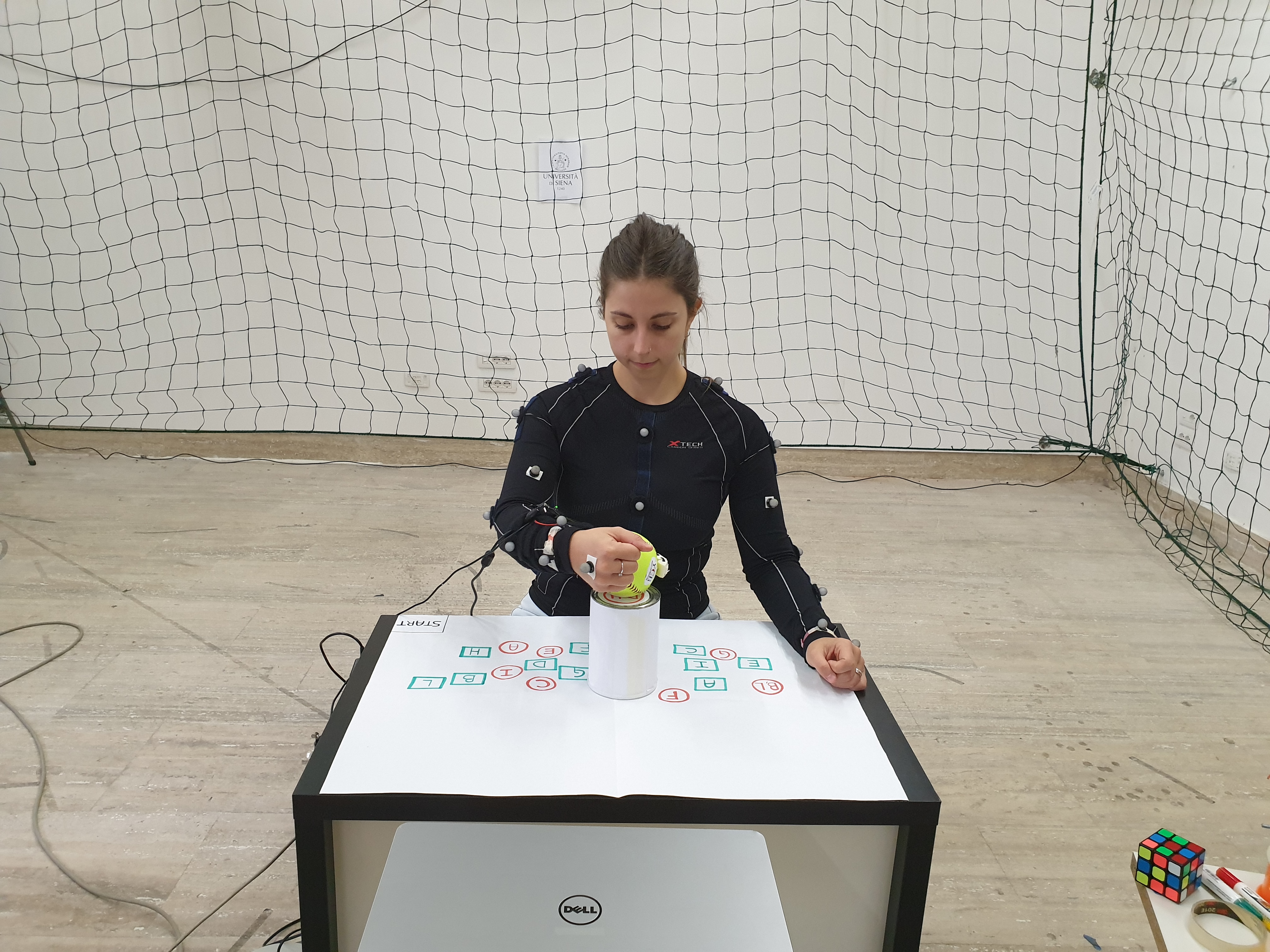}}
\hfill
\subfloat[]{\label{FIG:exp2_4}\includegraphics[trim={30cm 15cm  35cm 45cm}, clip,width=0.47\columnwidth]{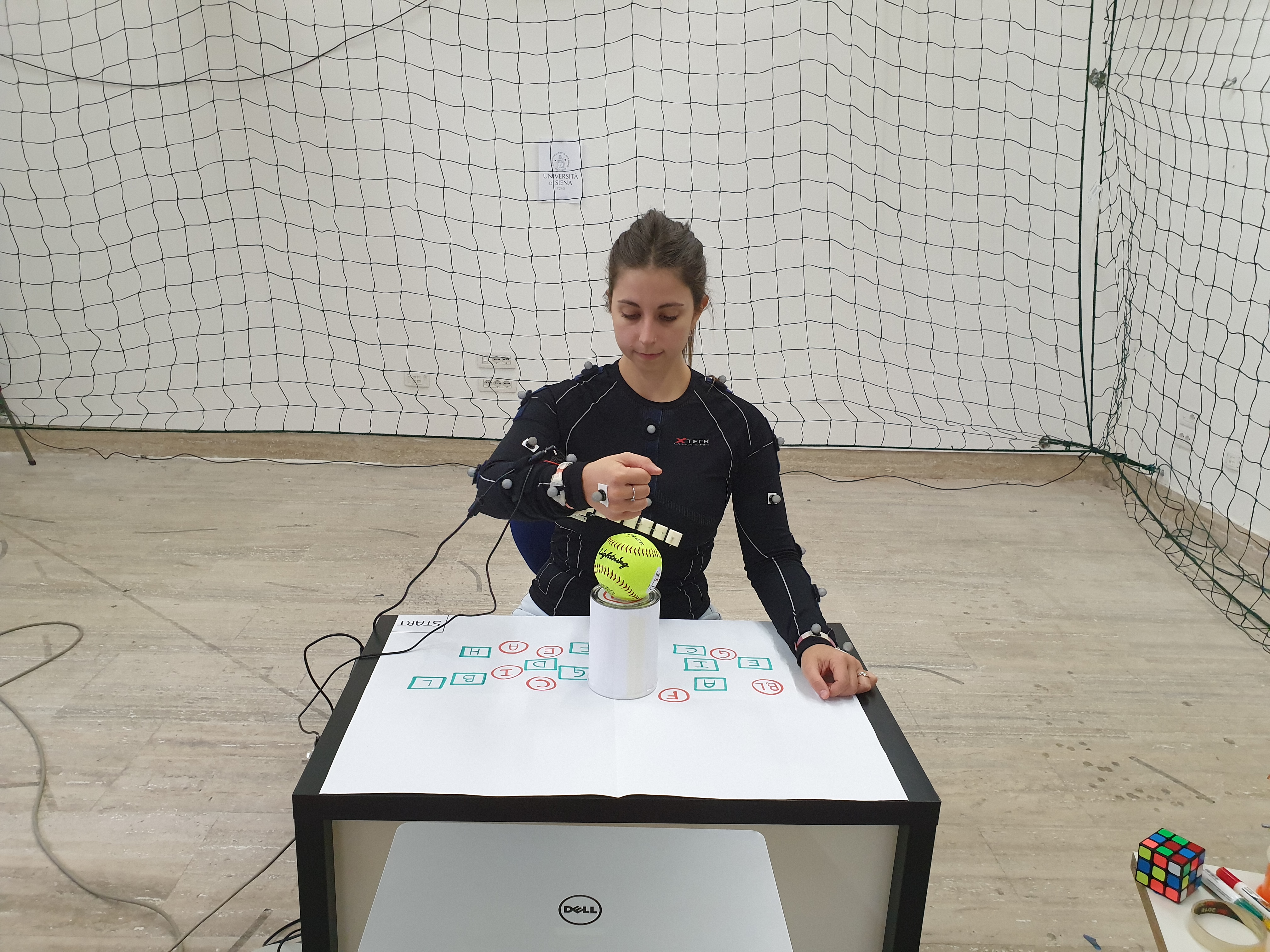}}
\hfill \null
\caption{Real Environment Experiment. Subjects were asked to pick (a), lift (b), place (c), and release (d) all the objects correctly, being as fast as possible and using only their right (impaired) arm. \label{FIG:exp2}}
\end{figure}

Subjects were seated in front of a table and asked to pick and place 10 objects taken from the YCB benchmark set \cite{calli2015ycb}. Details on the objects are reported in \tabb\ref{TAB:objects}. Objects were positioned on the table one at a time and each object was identified with an alphabetical code. 
The starting and target locations for the objects were marked on the table, %
with predefined positions: start positions were represented with a green square, whereas the goal positions were marked with a red circle. %
All the participants performed the same pick-and-place tasks. %
Participants were instructed to pick, lift, and place in the correct position as fast as possible each object, using only the right arm. Each pick-and-place task was considered successfully accomplished if the object was not dropped during the task execution and the elapsed time for the single pick-and-place task did not exceed \SI{20}{\s}.
A depiction of the scenarios is in \fig\ref{FIG:exp2}.

For the sake of comparison, we asked each subject to repeat the experiment using the supernumerary robotic finger controlled with a ring embedding a push button switch for opening/closing worn on the hand not involved in the task. This controller was chosen because it was considered the fastest and easiest control technique for operating the supernumerary robotic finger.

\begin{table}[t]

\centering
\begin{tabularx}{\columnwidth}{s|M|M|M}

\hline\hline 
\textbf{Code} &\textbf{Object} & \textbf{Weight [g]}  & \textbf{Dimensions [mm]} \\ \thickhline
A & Rubik's Cube & 84 & 60 x 60 x 60 \\ \hline
B & Soft Ball & 191 &96 \\ \hline
C & Chips Can & 205 &75 x 250 \\ \hline
D & Tomato Soup Can &  349 & 66 x 101 \\ \hline
E & Cups 1 & 14  & 60 x 62 \\ \hline
F & Cups 2 & 21  & 75 x 68 \\ \hline
G & Cups 3 & 28  & 85 x 72 \\ \hline
H & Cups 4 & 35  & 95 x 76 \\ \hline
I & Apple & 68 & 75 \\ \hline
L & Wine glass & 133& 89 x 137 \\ \hline
M & Potted Meat Can & 370 & 50 x 97 x 82
\\ \hline\hline
\end{tabularx}
\caption{Details on the objects used for the Real Environment Experiment. Objects are taken from the YCB benchmark set~\cite{calli2015ycb}. \label{TAB:objects}}
\end{table}

\pg{Metrics of interest}
Time to accomplish the task and number of successes and failures were considered as metrics for evaluating the performance of the participants.

\pg{Results}
As a first result, users took on average 8.94$\pm$3.26 \si{s} to accomplish the pick-and-place task when exploited the IKNS control (there were 12 failed attempts out of 100 trials which are not considered in the average accomplish time), while they needed on average 6.34$\pm$0.71 \si{\s} when they executed the task with the ring, regardless of the particular object. 
As expected, elapsed times in the latter case are lower for all the users and for all the objects. The reason why this is not a surprising result is that  people are generally familiar with controlling robotic devices using buttons. Moreover, the small standard deviation among the trials further highlights the predisposition of users in using this interface and confirms its effectiveness for comparison purposes. Differently, the IKNS strategy presents larger standard deviations. We interpreted this aspect as an indication of the fact that users need more time and more practice to gain further confidence and acquaintance with the system in real scenarios. %
Finally, it is worth noticing that %
the higher number of failures (4 failures out of 10 trials) was observed with object~I (\iec the apple), and it was reasonably due to the particular shape of the object. All the other objects have at most one failure.

To reinforce our hypothesis, we conducted a statistical analysis on the data. Times needed for moving each object with the ring and with the IKNS control were compared. Trials in which users failed were removed and not considered in the analysis. Data were neither normally distributed, nor symmetrical with respect to the median. Thus an exact sign test was used to compare the performance differences among the two trials. Outcomes of the test confirmed that performing the trials using the IKNS elicited a statistically significant median increase in time (\SI{1.44}{\s}) compared to the ring modality, $p < 0.01$.

\section{Conclusions and future work\label{SEC:concl}}
This study presented a new approach for controlling SRLs exploiting the redundancy of the human body. This kind of control takes advantage of movements in the Intrinsic Kinematic Null Space to enable the user to control supernumerary robotic limbs using the natural limb already involved in the task. Although IKNS control can be adopted to cooperate with SRLs in numerous scenarios (\eg surgical interventions, handling of loads, etc.), its potential becomes even more evident in the case of users with disabilities, as it overcomes the compromise which is often implicit in the standard \textit{non-autonomous} control: patients can recover part of their lost functionalities with SRLs, but the dexterity of their unimpaired limbs is reduced by the need to control the wearable robots. 

To provide an evaluation of the IKNS control, firstly we tested the framework in a virtual environment assessing the user capability of obtaining a desired control signal. Given the promising results, we asked subjects to test the IKNS control using a supernumerary robotic finger in the real environment in a representative ADLs task (such as picking, moving, and placing real objects). 

Results of the experimental campaign demonstrated that the proposed control strategy is suitable for controlling an additional DoF. Indeed, subjects showed good skills in cooperating with the supernumerary robotic finger in a relatively short time and little practice. 

Future research directions include quantifying how fast users control capabilities improves with practice, as well as expanding the IKNS approach to govern a larger number of degrees of freedom.

\bibliographystyle{IEEEtran}
\bibliography{biblio/conference,biblio/IEEEabrv,biblio/biblioAll}

\end{document}